\pdfoutput=1%Para compilar en arXiv
\PassOptionsToPackage{unicode}{hyperref}
\documentclass[acmsmall,screen,nonacm]{acmart}%Default
\AtBeginDocument{%
  }

\setcopyright{acmlicensed}
\copyrightyear{2026}
\acmYear{2026}
\acmDOI{XXXXXXX.XXXXXXX}

\acmJournal{TELO}

\acmVolume{}
\acmNumber{}
\acmArticle{}

\usepackage[ruled,linesnumbered]{algorithm2e}
\usepackage{float}
\usepackage{makecell, multirow}
\usepackage{color}

\definecolor{myyellow}{RGB}{204, 166, 48   }
\definecolor{mypurple}{RGB}{177, 8, 77    }
\definecolor{myblue}{RGB}{2, 14, 113}
\definecolor{mygreen}{RGB}{2, 113, 2}
\definecolor{mygrey}{RGB}{128, 128, 128  }

\RestyleAlgo{ruled}
\SetKwInput{KwInput}{Input}
\SetKwInput{KwOutput}{Output}

\DeclareMathOperator*{\argmax}{\arg\max}

\begin{document}

%%
%% The "title" command has an optional parameter,
%% allowing the author to define a "short title" to be used in page headers.

\title{A Review on Single-Problem Multi-Attempt Heuristic Optimization}

%%
%% The "author" command and its associated commands are used to define
%% the authors and their affiliations.
%% Of note is the shared affiliation of the first two authors, and the
%% "authornote" and "authornotemark" commands
%% used to denote shared contribution to the research.
\author{Judith Echevarrieta}
\affiliation{%
  \institution{BCAM - Basque Center for Applied Mathematics}
  \country{Spain}}
 \affiliation{%
 \institution{UPV/EHU - University of the Basque Country}
 \country{Spain}}
\email{judith.echevarrieta@ehu.eus}
\orcid{https://orcid.org/0000-0002-5674-889X}

\author{Etor Arza}
\affiliation{%
  \institution{NTNU - Norwegian University of Science and Technology}
  \country{Norway}}
  \email{etor.arza@ntnu.no}
\orcid{https://orcid.org/0000-0002-8044-0334}

\author{Aritz Pérez}
\affiliation{%
 \institution{BCAM - Basque Center for Applied Mathematics}
 \country{Spain}}
 \email{aperez@bcamath.org}
 \orcid{https://orcid.org/0000-0002-8128-1099}
 
\author{Josu Ceberio}
\affiliation{%
 \institution{UPV/EHU - University of the Basque Country}
 \country{Spain}}
 \email{josu.ceberio@ehu.eus}
 \orcid{https://orcid.org/0000-0001-7120-6338}
 
\thanks{
© [Judith Echevarrieta, Etor Arza, Aritz Pérez, Josu Ceberio] [2026]. This is the author's version of The Work. It is posted here for your personal use. Not for redistribution. The definitive version was published in ACM Transactions on Evolutionary Learning and Optimization, https://doi.org/10.1145/3795878. \bigskip}

%%
%% By default, the full list of authors will be used in the page
%% headers. Often, this list is too long, and will overlap
%% other information printed in the page headers. This command allows
%% the author to define a more concise list
%% of authors' names for this purpose.
%\renewcommand{\shortauthors}{Trovato et al.}

%%
%% The abstract is a short summary of the work to be presented in the
%% article.
\begin{abstract}

In certain real-world optimization scenarios, practitioners are not interested in solving multiple problems but rather in finding the best solution to a single, specific problem. When the computational budget is large relative to the cost of evaluating a candidate solution, multiple heuristic alternatives can be tried to solve the same given problem, each possibly with a different algorithm, parameter configuration, initialization, or stopping criterion. In this practically relevant setting, the sequential selection of which alternative to try next is crucial for efficiently identifying the best possible solution across multiple attempts. However, suitable sequential alternative selection strategies have traditionally been studied separately across different research topics and have not been the exclusive focus of any existing review. As a result, the state-of-the-art remains fragmented for practitioners interested in this setting, with surveys either covering only subsets of relevant strategies or including approaches that rely on assumptions that are not feasible for the single-problem case.

This work addresses the identified gap by providing a focused review of single-problem multi-attempt heuristic optimization. It brings together suitable strategies for this setting that have been studied separately through algorithm selection, parameter tuning, multi-start, and resource allocation. These strategies are described using a unified terminology within a common framework, which supports the construction of a taxonomy for systematically organizing and classifying them. The resulting comprehensive review facilitates both the identification and the development of strategies for the single-problem multi-attempt setting in practice.

\end{abstract}

%%
%% The code below is generated by the tool at http://dl.acm.org/ccs.cfm.
%% Please copy and paste the code instead of the example below.
%%


%%
%% Keywords. The author(s) should pick words that accurately describe
%% the work being presented. Separate the keywords with commas.

%\received{date}
%\received[revised]{date}
%\received[accepted]{date}

%%
%% This command processes the author and affiliation and title
%% information and builds the first part of the formatted document.
\maketitle

\newpage

\section{Introduction}\label{SECTION_Introduction}
% Unico problema/instancia
An optimization problem consists of finding the best solution from a search space that either maximizes or minimizes a given objective function, subject to specific constraints~\cite{arora_optimization_2015}. These problems are encountered in a wide range of disciplines, including logistics, scientific research, and engineering. In some cases, the objective is to solve multiple related problems, each defined by variations in the search space and the objective function. \citet{gago-carro_stochastic_2024} address a location-assignment problem to find the ambulance service management that maximizes the number of emergency calls attended under different demand scenarios. In other situations, the goal is to solve a single, well-defined problem. This setting is common in real-world applications where the interest lies in optimizing a specific search space and objective function. For example, in computational biology, \citet{grado_bayesian_2018} aim to identify the set of brain stimulation parameters that minimizes beta power in patients with Parkinson's disease. In renewable energy systems, \citet{reddy_wind_2020-1} searches for the wind farm layout that maximizes annual energy production based on a given terrain and wind conditions, while \citet{zarketa-astigarraga_holistic_2023} optimize the design of a Wells turbine to maximize its efficiency across a selected set of representative sea states. In this work, we focus on the second single-problem scenario.

% Metodos heuristicos
Heuristic and metaheuristic methods\footnote{Hereafter, for simplicity, we refer to both heuristic and metaheuristic methods as heuristic methods.} are commonly employed to solve such single optimization problems. In the previous examples, \citet{grado_bayesian_2018} apply an adaptive Bayesian optimization algorithm, while \citet{zarketa-astigarraga_holistic_2023} perform a genetic algorithm-based optimization. Although these problems are not always formally proven to be NP-hard, they are typically challenging to solve~\cite{kazimipour_towards_2019} due to characteristics such as the high dimensionality of large-scale solution spaces~\cite{mahdavi_metaheuristics_2015,zhan_survey_2022,nikeghbali_recent_2022}, non-convex and non-differentiable black-box objective functions~\cite{kazimipour_towards_2019,talbi_metaheuristics_2009,zhi-quan_luo_introduction_2006,pardalos_black_2021,sala_benchmarking_2020}, and noise in stochastic objective functions and constraints~\cite{fu_feature_2002-1,amaran_simulation_2016}. In such contexts, heuristic methods are particularly useful, as they provide acceptable solutions within reasonable computational times, even though they do not guarantee global optimality of the returned solution.

% Multiples-intentos
Heuristic methods usually rely on selecting and evaluating different solutions from the search space to return the one with the best objective value. In practice, when solving a single optimization problem, the evaluation of the objective function for each solution often takes only a few seconds~\cite{grado_bayesian_2018,reddy_wind_2020-1,zarketa-astigarraga_holistic_2023,cutler_real-world_2015,robinson_multifidelity_2006-1}. With such fast evaluations relative to the available computational budget, the practitioner has a large number of objective function evaluations at its disposal, enough to execute several heuristic methods until convergence. In such cases, the intelligent use of the abundant resources available is crucial to find the best possible solution to the problem of interest, as multiple attempts with different heuristic methods are possible. Nevertheless, the sequential selection of methods becomes a challenge.

% Ejemplificar el caos existente para resolver problema que resulta de unir las tres cosas anteriores (single-problem, multi-try, heuristic optimization)
There are different strategies in the literature that address the selection of the heuristic method to be considered in each attempt to solve a single optimization problem. Some works focus on restart strategies for a single local search algorithm~\cite{lopez-sanchez_multi-start_2014,mathesen_stochastic_2021}, which allow solving the same problem in multiple attempts, considering different initializations of the algorithm at each one. \citet{luersen_globalized_2004} focus on the kernel density estimation-based restart of the Nelder-Mead algorithm. \citet{auger_restart_2005} instead, implement a restart strategy for the Covariance Matrix Adaptation Evolution Strategy, where the initial population is resampled uniformly in the search space at each restart, and the population size is progressively increased following a predefined schedule. This allows executing each attempt not only with a newly sampled initial population, but also with a systematically varied population size parameter, which is known to significantly influence the behavior and performance of evolutionary algorithms. Other works consider the joint selection of multiple algorithms along with their configurations. \citet{souravlias_parallel_2019} address a combinatorial optimization problem by defining a strategy to distribute the available computational budget across multiple attempts, each corresponding to a different configuration of Tabu Search, Variable Neighbourhood Search, or Multistart Local Search algorithms. Finally, we identify successive halving-based strategies~\cite{li_hyperband_2018,das_dores_bandit-based_2018}, which primarily focus on defining stopping criteria for each attempt. These approaches aim to allocate more evaluations to parameter configurations of supervised learning algorithms that show greater potential, in order to converge toward the best-performing configuration that minimizes the empirical error.

% Hacer explicito el caos basandome en los ejemplos (las estrategias son muy diferentes, se estudian por separado y las review existentes no llegan a unificarlas todas)
The above examples show a wide diversity of strategies for addressing a single problem in multiple attempts. The diversity lies not only in the selection criteria, but also in the component that differentiates the heuristic methods selected for each attempt: a heuristic algorithm, a configuration of its parameters, an initialization or a stopping criterion. Thus, each combination of these four components defines a heuristic alternative to solve the optimization problem of interest. We have observed that alternative selection strategies focus on solving a single problem are divided into four different topics studied separately in the literature, each one focused mainly on the selection of one component. However, all these strategies are valid for the same practical \textit{Single-Problem Multi-attempt Heuristic Optimization} (SIMHO) scenario, where a practitioner faces the following question: how to sequentially select heuristic alternatives to find the best possible solution to a given optimization problem in multiple attempts?. There are reviews that give an overview of alternative selection strategies~\cite{bessiere_algorithm_2016,gendreau_intelligent_2019}, but most either focus on limited subsets of relevant strategies for SIMHO~\cite{locatelli_global_2016,gendreau_intelligent_2019,yang_hyperparameter_2020} or include approaches that rely on assumptions that make them not feasible or not efficiently applicable to the SIMHO setting~\cite{huang_survey_2020,kerschke_automated_2019,bischl_hyperparameter_2023}.

Existing reviews do not comprehensively cover strategies for practitioners facing specific SIMHO settings. To overcome this gap, this paper presents a review focused on selection strategies suitable for SIMHO, encompassing strategies that address the selection of any combination of the four components. We unify existing literature related to SIMHO within a single framework, identifying a shared structure among strategies typically studied separately and formalizing them under the same terminology. Furthermore, we define a taxonomy that characterizes all strategies within the proposed framework. To the best of our knowledge, this is the first review to unify, describe, and classify selection strategies exclusively for SIMHO while considering all four components together.

% Organizacion
The rest of the paper is organized as follows. Section~\ref{SECTION_RelatedWork} classifies related alternative selection strategies, identifies in that classification those valid for SIMHO, and highlights the difference of the review proposed in this work from the perspective that already exists. Section~\ref{SECTION_Framework} introduces the framework that sets the notation, formalizes SIMHO as an alternative selection problem, and defines the abstraction that generalizes within the same terminology the existing suitable strategies. Section~\ref{SECTION_RelatedTopics} describes the four topics in the literature about alternative selection that address SIMHO, illustrating how all of them fit into the proposed abstraction. Section~\ref{SECTION_Taxonomy_UpdateDistribution} defines the taxonomy and characterizes the different strategies as particular cases of the framework. Section~\ref{SECTION_Discussion} discusses the utility and scope of the proposed framework-based review. Finally, Section~\ref{SECTION_Conclusion} concludes the paper.

\section{Related work}\label{SECTION_RelatedWork}
There are several selection strategies aimed at identifying the best heuristic alternative from an available set. We distinguish two groups of selection strategies based on the type of information they rely on. To address a single optimization problem, \textit{single-problem} strategies use information gathered during the optimization process of that problem. In contrast, \textit{multi-problem} strategies extract knowledge from a set of previously solved problems and transfer it to a new problem. In multi-problem strategies, both the problems used to extract information and the target problem usually belong to a family of related problems, where each problem is referred to as an \textit{instance}. 

\subsection{Single- and multi-problem strategies through the lens of SIMHO}\label{SECTION_SingleProblemMultiProblem}
% Multi-problme (ejemplos + limitaciones)
Examples of multi-problem strategies include per-instance algorithm selection~\cite{kerschke_automated_2019,rice_algorithm_1976} and offline parameter tuning~\cite{hamadi_automated_2011,coello_sequential_2011-1}, in which the selection is based on a supervised model trained using features extracted from a set of problem instances and the observed performance of a set of algorithms or configurations on them. Given the features of a new instance, the model recommends the most appropriate alternative. The Irace package~\cite{lopez-ibanez_irace_2016} provides selection strategies that replace supervised models with statistical racing procedures~\cite{maron_racing_1997,birattari_racing_2002}, aiming to identify algorithm configurations that perform robustly across a set of instances. Iteratively, a set of configurations is sampled from a probabilistic model and evaluated on a representative subset of instances. The best-performing configurations are selected to update the model, progressively biasing the sampling process toward more promising regions of the configuration space.

A major limitation of multi-problem strategies is that they must invest a significant number of attempts to build accurate models by solving multiple related instances prior to addressing the target problem, which often results in a time-consuming process~\cite{lobo_parameter_2007}. Furthermore, their effectiveness depends on access to a sufficiently large and representative set of instances~\cite{bartz-beielstein_online_2014,kerschke_automated_2019}, which is crucial for ensuring generalization to unseen problems. This may not be feasible in settings such as SIMHO, where only a specific optimization instance is available or the focus. Nevertheless, multi-problem strategies remain highly valuable in contexts where users must routinely address problems belonging to the same family~\cite{roman_bayesian_2016}.

% Hyper-heuristics (tanto single-problem como multi-problem + limitaciones)
Hyper-heuristics~\cite{di_chio_investigation_2011,drake_recent_2020} are high-level strategies designed to automatically generate an effective heuristic algorithm for solving a specific problem during a single run. They achieve this by chaining low-level heuristics, such as swap or shift neighborhood moves in iterative local search~\cite{burke_iterated_2010,kanade_parameter-less_2004}, or crossover and mutation operators of evolutionary algorithms, as used in parameter control~\cite{karafotias_parameter_2015} and adaptive operator selection~\cite{fialho_analyzing_2010-1}. We consider these hyper-heuristics to fall within the single-problem group, as they adapt to the particular characteristics of the target problem, considering only the information collected during its optimization process. These strategies are typically defined as an iterative process composed of three key tasks: 1) credit assignment, which scores the performance of each low-level heuristic~\cite{fialho_toward_2010}; 2) a selection mechanism, which chooses the next heuristic to apply based on these scores; and 3) a move acceptance to determine whether the new solution should be accepted. \citet{correia_switch_2023} propose a per-run trajectory-based algorithm-switching policy that relies exclusively on local landscape features observed during the optimization of the target problem. Their results demonstrate that switching between two evolutionary algorithms within a single optimization process can significantly improve performance by leveraging the strengths of each algorithm. However, there also exist multi-problem hyper-heuristics, where the selection mechanism is based on the information gathered from previously solved problems. For example, \citet{arza_adaptive_2020} introduce a population-based hyper-heuristic that trains a neural network on an instance and then uses the trained model to guide the selection, and application of low-level operators across other instances of the same problem family. 

The performance of hyper-heuristics mainly depends on the selection mechanism~\cite{scoczynski_selection_2021,bouneffouf_survey_2020,sutton_reinforcement_2018,li_adaptive_2014,dacosta_adaptive_2008} and the parameters on which all three tasks may depend~\cite{doerr_understanding_2020}. An appropriate tuning is often necessary to obtain high-quality solutions for a specific problem~\cite{ferreira_multi-armed_2017,maturana_extreme_2009}. For instance, \citet{vrugt_self-adaptive_2009} define a population-based hyper-heuristic where the population size parameter is incrementally adjusted for the problem being addressed. In the SIMHO scenario, as multiple attempts are available to solve a single optimization problem, we consider single-problem hyper-heuristics as another alternative to select from for valid SIMHO strategies, as they can be used together with their parameter configuration, initialization, and stopping criterion to carry out a specific attempt.

\subsection{Existing reviews through the lens of SIMHO}

% Concretar la diferencia de nuestra review con las existentes despues de haber hablado sobre los diferentes topics relacionados
Overall, each suitable strategy for SIMHO may focus on the selection of different components. There are multiple reviews in the literature that include single-problem selection strategies valid for SIMHO. However, they typically cover at most the selection of two or three out of the four components (see the third column of Table~\ref{TABLE_contribution}). This is because, in most cases, their primary aim is not to unify independently studied strategies across the different topics relevant to SIMHO. Instead, they provide broader overviews of alternative selection strategies that often include related but misaligned approaches, such as multi-problem strategies that do not apply to the SIMHO setting (see the second column of Table~\ref{TABLE_contribution}). 

For instance, \citet{souravlias_algorithm_2021} include algorithm portfolios and per-instance algorithm selection, \citet{bessiere_algorithm_2016} differentiates between strategies based on dynamic and static features, \citet{huang_survey_2020} classify strategies as specialist or generalist, \citet{schede_survey_2022} refer to instance-specific and offline configurators, \citet{elshawi_automated_2019} describe both hyperparameter optimization and meta-learning strategies, or \citet{bischl_hyperparameter_2023} which briefly discuss warm-start strategies for reducing computational time in hyperparameter optimization by transferring information from previous experiments on the problem of interest or similar problems to the one currently being considered. In all these cases, their dual focus fits our definitions of single-problem and multi-problem strategies, respectively. However, we compare our work with these reviews only through their parts dedicated to single-problem strategies, as the remaining parts fall outside the scope of SIMHO.

\begin{table}[ht]
\centering
\caption{Classification of existing reviews according to the type of selection strategies they focus on and the components addressed that are valid for SIMHO.}\label{TABLE_contribution}

{\footnotesize
\begin{tabular}{|c|cc|cccc|}
\hline 
\multirow{3}{*}{Reviews} & \multicolumn{2}{c|}{\makebox[3.1cm]{Focus on}} &\multicolumn{4}{c|}{Components} \\
& {\footnotesize \makebox[1.55cm]{Single-problem}} & {\footnotesize \makebox[1.55cm]{Multi-problem}}&{\footnotesize \thead{Heuristic \\algorithm}}& {\footnotesize \thead{Parameter \\configuration}} & {\footnotesize Initialization} & {\footnotesize \thead{Stopping \\criterion}} \\
\hline  
{\footnotesize\thead{\citep{souravlias_algorithm_2021}}} & \checkmark & \checkmark & \checkmark & \checkmark & &  \\ 
\hline 
{\footnotesize\thead{\citep{bessiere_algorithm_2016}}} & \checkmark & \checkmark & \checkmark & \checkmark & \checkmark&  \\
\hline 
{\footnotesize\thead{\citep{huang_survey_2020}\\\citep{schede_survey_2022}}}  & \checkmark & \checkmark &  & \checkmark & &  \\ 
\hline 
{\footnotesize\thead{\citep{locatelli_global_2016}\\\citep{gendreau_intelligent_2019}}} & \checkmark & & & & \checkmark & \\ 
\hline 
{\footnotesize\thead{\citep{yang_hyperparameter_2020}}} & \checkmark & & & \checkmark &  & \checkmark  \\ 
\hline 
{\footnotesize\thead{\citep{elshawi_automated_2019}}} & \checkmark & \checkmark & & \checkmark &  & \checkmark  \\ 
\hline 
{\footnotesize\thead{\citep{bischl_hyperparameter_2023}}} & \checkmark & \checkmark & & \checkmark &  \checkmark & \checkmark  \\
\hline 
This work & \checkmark & & \checkmark & \checkmark &\checkmark &\checkmark \\ 
\hline 
\end{tabular} }
\end{table}

As a result of the dual focus of existing reviews, they often lack a unified formalization and rely on fragmented taxonomies with component-specific descriptions. In contrast, we provide a review that focuses exclusively on single-problem multi-attempt strategies, and we introduce a common framework that formalizes the strategies that address the selection of any combination of the four components using consistent terminology. This unified perspective enables a more coherent and comprehensive taxonomy, providing a clearer and more practical reference for researchers and practitioners working specifically on SIMHO, while avoiding the inclusion of strategies that are not efficiently applicable.

\section{SIMHO framework}\label{SECTION_Framework}
An \textit{optimization problem} consists of finding the best solution among a set of possible ones, which maximizes a performance function. Formally, it could be defined as follows
\begin{equation}\label{EQUATION_OptimizaionProblem}
\argmax_{x\in X}f(x)
\end{equation}
where $X$ is the \textit{solution space} and $f$ is the \textit{objective function} that measures how good a solution is\footnote{Without loss of generality, we assume that the optimization problem is a maximization problem. In the case of minimization, it would be enough to replace $f(x)$ in Problem~(1) with $-f(x)$. In addition, as both the solution space $X$ and the objective function $f$ are specific, Problem~(1) defines a single, well-defined optimization problem.}. In this section, we define the conditions considered to solve Problem~(1), and the generalization of the suitable literature strategies to address it.

\subsection{Multi-attempt heuristic optimization}
Certain real-world problems are best solved using heuristic optimization methods. \textit{Heuristic algorithms} are optimization algorithms that provide acceptable solutions in reasonable computational times, without any guarantee of optimality. A wide variety of such algorithms have been developed over the last decades~\cite{vermetten_large-scale_2024}, including direct search methods (e.g., Nelder–Mead~\cite{nelder_simplex_1965} and Hooke-Jeeves~\cite{hooke_direct_1961} algorithms), evolutionary algorithms (e.g., Genetic Algorithms~\cite{glover_handbook_2003}, Estimation of Distribution Algorithms~\cite{larranaga_review_2002} and population-based hyper-heuristics~\cite{yuen_which_2013,fei_peng_population-based_2010}) and gradient-based methods (e.g., RMSProp~\cite{tieleman_lecture_2012} and ADAM~\cite{kingma_adam_2014} considered when training large neural networks). With sufficiently large computational budgets, multiple executions of different heuristics can be afforded, each providing an \textit{attempt} to solve the problem. 

To carry out each available attempt, four \textit{components} must be fixed: 1) the heuristic algorithm, 2) its parameter configuration, 3) its initialization and 4) the stopping criterion that limits its execution time. A combination of these four components represents a reproducible sequence of computational instructions~\cite{arza_fair_2024}, which we call \textit{alternative}. Its execution returns the best-visited solution in terms of the objective function. The objective value of this solution determines the \textit{quality} of the alternative, which can only be revealed by executing it. Formally, we define the set of alternatives as
\begin{equation}\label{DEFINITION_A}
A=\big\lbrace (h,\theta,\xi,c)~|~h\in H,\theta\in \Theta, \xi\in \Xi,c\in C\big\rbrace
\end{equation}
where $H$, $\Theta$, $\Xi$ and $C$ are the sets of possible heuristic algorithms, parameter configurations, initializations and stopping criteria, respectively (see Appendix~\ref{APPENDIX_HierarchicalDep} for a more detailed description). The contribution of each alternative $a\in A$ to the optimization is given by the \textit{information} it provides after its execution, which in practice is limited to the following triplet
\begin{equation}\label{DEFINITION_information}
I_a=\lbrace x_a,t_a,f_a\rbrace
\end{equation}
where $x_a$ is its best-visited solution, $t_a$ is its total execution time and $f_a=f(x_a)$ is its quality. 

In the described context, we define the \textit{multi-attempt heuristic optimization} as the search for the solution to Problem~\eqref{EQUATION_OptimizaionProblem} with the highest possible objective value under the following conditions:
\begin{itemize}
\item[•] \textit{Availability of different alternatives}. A set of alternatives with a priori unknown qualities is available\footnote{In this work, we consider the set of alternatives as given and its definition is out of the scope of this paper. For common challenges and tips on defining an appropriate set of alternatives see~\cite{souravlias_algorithm_2021,bessiere_algorithm_2016}.}.
\item[•] \textit{Availability of multiple attempts}. The computational budget is large enough such that different alternatives can be executed.
\end{itemize}
As the quality of each alternative is the objective value of the best-visited solution, the single-problem multi-attempt heuristic optimization can be reformulated as an alternative selection problem as follows
\begin{equation}\label{EQUATION_AOproblem}
\argmax_{a\in A} f_a
\end{equation}
where the alternative of the highest possible quality must be found over the multiple available executions\footnote{Despite the formal similarity to other existing problems (e.g., the classic algorithm selection introduced by~\cite{rice_algorithm_1976} for multi-problem setting) this formulation is specific to single-problem settings because it depends on a single optimization problem determined by Problem~\eqref{EQUATION_OptimizaionProblem}. For a more detailed discussion, see Appendix~\ref{APPENDIX_MultiproblemSetting}.}. Thus, the reformulation shifts the exploration of the solution space $X$ to the set of alternatives $A$. From this point of view, finding the best $a\in A$ that solves Problem~\eqref{EQUATION_AOproblem} becomes the new optimization problem, where the desired solution is the best-visited solution $x_a$ provided by the $a \in A$ with the highest quality $f_a$. We define this problem generically and any approach that addresses it as SIMHO problem and strategy, respectively.

\subsection{The abstract \mbox{SIMHO} strategy}\label{SECTION_Algorithm}
A naive \mbox{SIMHO} strategy consists of iteratively executing a different alternative $a\in A$ drawn from a uniform random distribution to reveal its quality until the available executions are exhausted, and thus keeping the highest quality solution among all of them. However, this randomness does not consider the information observed in the previous executions, which may result in the selection of similar alternatives that have already shown poor quality~\cite{mathesen_stochastic_2021}. In the efficient \mbox{SIMHO}~strategies identified in different topics of alternative selection, it is crucial the use of the information from previously executed alternatives to guide the selection of the remaining executions. In other words, the information observed from executed alternatives is used to define the selection preference of a new alternative. 

Let $\mathcal{I}$ be the set formed by the information of executed alternatives, which we call \textit{observed information}. We define the \textit{probability} of an alternative \mbox{$a=(h,\theta,\xi,c)\in A$} as a factorization of conditional distributions that determine the preference of each component to be selected for the next attempt conditioned on those selected before it
\begin{equation}\label{DEFINITION_Probability}
p(h,\theta,\xi,c)=p_H(h)\cdot p_\Theta(\theta\mid h)\cdot p_\Xi(\xi\mid h, \theta)\cdot p_C(c\mid h, \theta, \xi)
\end{equation}
where $p$, $p_H$, $p_\Theta$, $p_\Xi$ and $p_C$ denote the probability distributions over the set of alternatives $A$ and the sets of components $H$, $\Theta$, $\Xi$ and $C$, which we call \textit{distribution of alternatives} and \textit{component distributions}, respectively. Notice that to define the probability of $a$, its factors could use the informations from $\mathcal{I}$ associated with the executed alternatives with the same heuristic algorithm, or the same algorithm and parameter configuration, or the same algorithm, configuration and initialization (see Appendix~\ref{APPENDIX_HierarchicalDep} for a more detailed description).

We propose an abstract definition that formalizes the existing \mbox{SIMHO} strategies proposed in separately studied topics under the same terminology. This consists of an iterative alternation of two processes: the \textit{information generation} and the \textit{distribution update}. The first focuses on expanding the observed information $\mathcal{I}$ to enable the second to update the distribution of alternatives $p$ based on it. The pseudocode associated with our proposal is shown in Algorithm~\ref{ALGORITHM}. It starts by initializing $p$ (line~2). Then, until the available budget is exhausted, in each iteration the alternation of the two processes is applied. Firstly, to generate new information, an unexecuted alternative $a$ is sampled from $A$ based on its current distribution (line~4). Then, the selected alternative is executed to reveal its information $I_a$ (line~5) and add this new information to the observed information $\mathcal{I}$ (line~6). Secondly, the distribution is updated by reassigning to each unexecuted alternative a probability based on the observed information (line~7). Finally, when no more time is left to carry out a new iteration, the best-visited solution with the highest objective value among all the executed alternatives is returned (line~10)\footnote{Abusing the notation, we write $x_a\in\mathcal{I}$ to refer solely to the third elements (best-visited solutions) of the triplets~\eqref{DEFINITION_information} that form the observed information $\mathcal{I}$.
In Sections~\ref{SECTION_ProbabilityUpdateEMQ} and~\ref{SECTION_ProbabilityUpdateEQI}, the same notation is used to refer solely to the observed execution times ($t_a\in\mathcal{I}$) and the observed qualities ($f_a\in\mathcal{I}$).
}. 

\begin{algorithm}[h]
{\small
\SetAlgoLined
\caption{Abstract \mbox{SIMHO} strategy}\label{ALGORITHM}
\KwInput{\\
~~$X,f\colon$ Solution space and objective function;\\
~~$A\colon$ Set of alternatives;\\
~~$t_{max}\colon$ Computational budget that allows multiple attempts.\\

\vspace{2pt}
\KwOutput{$\widetilde{x}\colon$ Best solution found.}
$\mathcal{I}\gets \emptyset$, $t\gets 0$ \;
$p\gets$ Initialize the distribution of $A$\;
\While{$t< t_{max}$}{
	$a\gets$ Sample $p$\;
	$I_a=\{x_a,t_a,f_a\}\gets$ Execute $a$ \;
	$\mathcal{I}\gets \mathcal{I}\cup\{ I_a\}$ \;
	$p\gets$ Update probability distribution over $A$ using $\mathcal{I}$\;
	$t\gets t +t_a$ \;
}
\Return{$\widetilde{x}=\argmax_{x_a\in \mathcal{I}} f_a$}
}
}
\end{algorithm}

\section{Topics addressing SIMHO}\label{SECTION_RelatedTopics}
We have observed that the strategies that address SIMHO are proposed within four different topics related to alternative selection. All of them address specific optimization problems in multiple attempts with a set of alternatives, and the strategies proposed in them focus on determining the alternative to be executed in each attempt until the available executions are exhausted. 
Specifically, each topic names the iterative process\footnote{The same or similar terms are used in the literature to refer to different topics that focus on different objectives, as has been shown throughout Section~\ref{SECTION_RelatedWork}, and has been specified in the second column of Table~\ref{TABLE_contribution}. 
It should be emphasized that, although we tried to choose the terms that best represent the single-problem strategies addressed by each topic, these terms have other uses in the literature.} depending on which of the four components of the alternative it determines. 
To illustrate how the proposed abstraction generalizes the existing strategies in the four topics, we describe each of them both in general terms and using illustrative examples within Algorithm~\ref{ALGORITHM}. For this purpose, we identify in them: an optimization problem, a set of alternatives, multiple attempts and the two alternating processes.

\subsection{Algorithm Selection} 
\textit{Algorithm Selection} (AS), also identified as \textit{Algorithm Portfolio} by~\citet{souravlias_algorithm_2021}, is motivated by the efficient allocation of computational resources between a set of heuristic algorithms that perform differently on the problem of interest. Consequently, AS iteratively specifies the heuristic algorithm in which to spend more resources.

An instantiation of our abstraction in the AS topic is provided by \citet{souravlias_parallel_2019} to solve the combinatorial problem called the detection of circulant weighing matrices~\cite{arasu_circulant_2010}, using a portfolio consisting of two variants of the algorithm Tabu Search~\cite{du_tabu_1998}, two variants of Variable Neighbourhood Search~\cite{mladenovic_variable_1997}, and three variants of Multistart Local Search~\cite{gendreau_iterated_2019}, under different parameter configurations. In each iteration, the number of available CPUs is distributed among the portfolio constituents proportionally to their performances in previous executions, to execute again in parallel new initializations of each one for the same maximum number of evaluations. Consequently, in this case, the algorithm, parameter configuration and initialization are the components that differentiate the alternatives. Moreover, running different times multiple algorithms also involves the availability of multiple attempts. Finally, the new qualities observed when running new initializations of each algorithm and configuration pair allow updating the estimated performance for each pair and proportionally associating the number of new initializations to execute next.

\subsection{Parameter Tuning} 
\textit{Parameter Tuning} (PT), denoted as \textit{specialist} PT by~\citet{huang_survey_2020}, focuses on automating the parameter setting of heuristic algorithms, to facilitate non-expert users in identifying well-performing configurations to the problem of interest and avoiding costly brute force techniques, such as grid search. This topic is closely related to evolutionary computation~\cite{larranaga_review_2002,glover_handbook_2003}, where Genetic Algorithms or Estimation of Distribution Algorithms, although originally proposed as efficient, easy to use and applicable to a wide range of problems~\cite{holland_adaptation_2010}, their effectiveness is extremely dependent on the correct selection of parameters such as population size, crossover probabilities and selection ratios~\cite{harik_parameter-less_1999}. This requires a tuning process, in which different configurations are tested to find the most appropriate one for the problem of interest. In machine learning, the topic is known as \textit{Hyperparameter Optimization} (HPO)~\citep{bischl_hyperparameter_2023} where the objective is to find the hyperparameter configuration of the learning algorithm of a model, which after being learned on a training data set has the minimum loss function value on a test data set. We do not know a priori which hyperparameter configuration will perform best, and thus the computational resources must be distributed among several learning executions carried out with different configurations. In short, PT determines in which parameter configuration to invest resources at each iteration.

\citet{doerr_understanding_2020} contribute to the design of parameter-less Estimation of Distribution Algorithms within the PT topic, proposing a smart restart strategy for the compact Genetic Algorithm (cGA)~\cite{harik_compact_1999} with exponentially increasing population size. Different population sizes of the algorithm are considered to solve a binary optimization problem. They start executing the algorithm with a small population size, which is stopped once the risk of genetic drift~\cite{doerr_sharp_2020} is deemed too high, to reinitialize the algorithm again from a uniform distribution but now associated with a larger population size. Thus, although the algorithm and stopping criterion are the same across all runs, initialization and configuration are the components defining the set of alternatives. Furthermore, the multiple restarts of the cGA with different population sizes lead to multiple attempts. Finally, in this case, the schedule to select the algorithm configuration for each execution is known in advance, since without having the information of each execution, the rule with which the population size will be updated is scheduled beforehand.

\subsection{Multi-Start} 
\textit{Multi-Start} (MS)~\citep{gendreau_intelligent_2019,locatelli_global_2016} is a technique used to improve the performance of heuristic algorithms, especially in non-convex problems where multiple local optima may exist. It alternates the execution of a single heuristic algorithm with its restart at another location in the solution space. Each initialization defines an independent optimization process, potentially leading to different local optima. In machine learning, HPO may require substantial computational resources, and one way to reduce the computational time are \textit{warm-starts}, discussed by~\citep{bischl_hyperparameter_2023} as MS strategies that use information from previous experiments to initialize new ones. Therefore, MS defines the new location to be considered, which determines the unknown performance of the algorithm until it is executed again.

For example, \citet{luersen_globalized_2004} solve a composite laminate design problem by considering only the Nelder-Mead algorithm~\cite{nelder_simplex_1965} with a fixed parameter configuration. It is restarted from a different location when the stopping criterion of reaching a small, flat or degenerate simplex in one iteration of the local search is satisfied. After each run of the algorithm, knowledge of the landscape and its local optima is enriched with the best-observed solution. This is used to update a probability distribution over the solution space, which determines the probability that the algorithm from each solution can explore a new local optimum. Thus, the component that determines the set of alternatives is the initialization, running different local searches implicitly assumes the availability of multiple attempts, and the best solution from each local search generates new information that updates the probability distribution over new initializations.

\subsection{Resource Allocation} 
\textit{Resource Allocation} (RA) focuses on speeding up the evaluation of a parameter configuration or initialization of a heuristic algorithm. This is done by gradually allocating more resources to the most promising configurations or initializations and discarding the poor ones early. When different hyperparameter configurations of the same learning algorithm are evaluated, this approach is also identified as \textit{Multi-fidelity} for HPO~\cite{bischl_hyperparameter_2023,elshawi_automated_2019,yang_hyperparameter_2020}. 
A monotonically increasing computational cost is considered for the model fitting, identifying higher fidelity evaluations with longer execution times in fits. The fidelity can be measured in terms of training epoch, number of iterations or train set size used in the previous ones for the learning algorithm. The last case carries out a \textit{Progressive sampling}~\cite{provost_efficient_1999}, where a sample of the entire available train set is considered and its size is progressively increased to retrain learning algorithms that have shown better results in previous trainings with smaller samples. In contrast, when evaluating different initializations of the same algorithm, the approach is related to MS and the resources are measured by the number of visited solutions. Therefore, dynamically allocating resources allows identifying unpromising alternatives quickly to have as much time as possible for those that seem to be promising according to the observed results. Thus, RA involves determining the amount of resources to be considered in each iteration. 

An illustrative example of a strategy in the RA topic is the Hyperband algorithm~\cite{li_hyperband_2018} designed to optimize the hyperparameters of a convolutional neural network. Hyperband simultaneously learns the weights of the neural network with different hyperparameter configurations for a few epochs at each attempt. After each batch, in the next attempt, it resumes the learning of the configurations that have led to the most promising networks for a few more epochs, and the process is repeated until only the network trained with the most promising hyperparameter configuration remains. Therefore, the optimization problem in this case consists of finding the weights of the neural network that minimize the loss function. For this purpose, different configurations of the parameters of the learning algorithm and the stopping criteria based on successive halving are considered, which define the set of alternatives. Finally, with the implicit assumption of the availability of multiple attempts, the different configurations are run with different incremental stopping criteria, to decide which configurations to continue running considering the observed performances.

\section{Taxonomy of SIMHO strategies}\label{SECTION_Taxonomy_UpdateDistribution}
\begin{figure}[h]
\centering
\includegraphics[width=12.5cm]{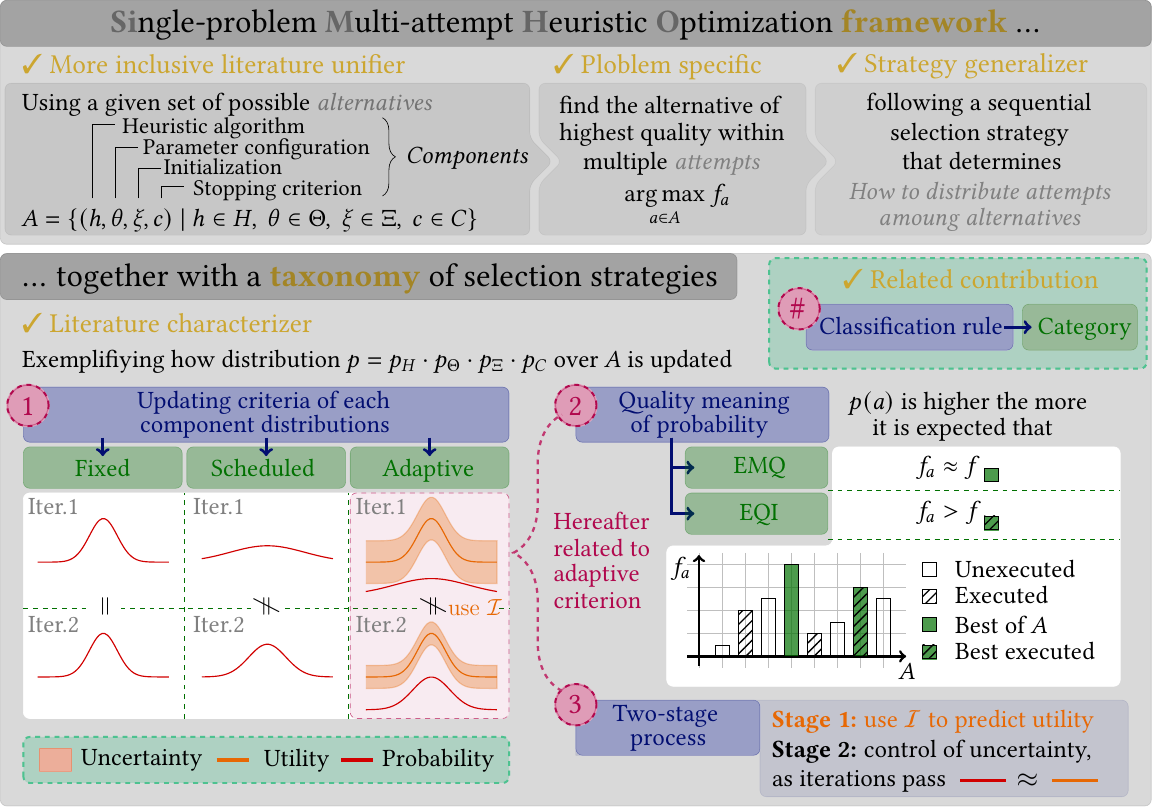}
\caption{The first box (named \textit{Single-problem Multi-attempt Heuristic Optimization framework...}) highlights in {\color{myyellow}\textbf{yellow}} the contributions of a review carried out throughout the proposed framework. The second box under it (named \textit{...together with a taxonomy of selection strategies}) enumerates in {\color{mypurple}\textbf{purple}} the linked classification rules in {\color{myblue}\textbf{blue}} of a \mbox{SIMHO} strategy, with the categories defined for each of them in {\color{mygreen}\textbf{green}}, and the contribution implied by the resulting taxonomy in {\color{myyellow}\textbf{yellow}}.}\label{FIGURE_summary}
\end{figure}

Addressing a literature revision under the proposed framework would lead to a review that: unifies the literature more inclusively, since it considers any combination of the four components as an alternative (see the sub box on the left inside the first box of Figure~\ref{FIGURE_summary}); is specific to researchers interested in solving a single optimization problem heuristically within multiple attempts (see the middle sub box in the first box of Figure~\ref{FIGURE_summary}); and summarizes the existing literature strategies from a shared terminology and perspective, avoiding individualistic descriptions (see the sub box on the right inside the first box of Figure~\ref{FIGURE_summary}). This section focuses on completing these three outcomes. Hence, hereafter we refer to and identify any strategy in the literature that addresses SIMHO equally from our Algorithm~\ref{ALGORITHM}. 

The aim of a \mbox{SIMHO} strategy is to determine the distribution of attempts among the given possible alternatives. Therefore, its effectiveness mainly depends on the distribution of alternatives defined at each iteration, which is the principal difference between its possible variants. Remember that the objective of the \mbox{SIMHO} problem is to find the alternative with the highest quality, since this is the one that returns the best-visited solution with the highest objective value. This alternative can only be selected by the sampling, which is guided by the distribution. We have observed that in the literature, its update is carried out as follows:

\begin{itemize}
\item First, it is determined how the probability distributions of the components will evolve with the observed information over the iterations. It should be considered which ones will remain unchanged, and which ones will be modified (see {\color{mypurple}\textit{Rule 1}} in the second box of Figure~\ref{FIGURE_summary}).
\item Then, for the distributions that are modified according to the observed information to guide the selection, the meaning that the probabilities will take to prioritize the selection is determined. A higher probability is likely to indicate a better alternative or an alternative closer to the optimum (see {\color{mypurple}\textit{Rule 2}} in the second box of Figure~\ref{FIGURE_summary}).
\item Finally, it is determined how to map the probabilities by making estimates from the observed information and controlling the uncertainty (see {\color{mypurple}\textit{Rule 3 }}in the second box of Figure~\ref{FIGURE_summary}).
\end{itemize}

In the following, we define a taxonomy with three linked classification rules associated with the three previous items. Each section defines each rule in order. In them, we determine the categories considered, illustrate each category with examples, and finally classify all the reviewed works accordingly. Overall, this section provides a review that characterizes existing single-problem selection strategies abstracting from the component. In fact, we will show how the same selection strategies sometimes appear in different topics to select different components. The papers included in this section extend and complete the illustrative examples described in Section~\ref{SECTION_RelatedTopics}, all of them presenting strategies that we consider single-problem\footnote{Some of the strategies reviewed in Section~\ref{SECTION_Taxonomy_UpdateDistribution} are presented in the literature without explicitly defining the specific solution space and objective function of the optimization problem they aim to address.
Nevertheless, these strategies remain inherently single-problem, as they rely exclusively on the information generated during the optimization of the same problem.}. These strategies fit with the abstract Algorithm~\ref{ALGORITHM} that formalizes under the same terminology those valid for SIMHO. They constitute the core of our review\footnote{The related reviews summarized in Table~\ref{TABLE_contribution} are not classified under our taxonomy, as they are used only for comparison purposes and are not part of the reviewed works themselves.}, which are summarized and classified through Tables~\ref{TABLE_Overview}, \ref{TABLE_UtilityPredictor}, and \ref{TABLE_UncertaintyControl}.

\subsection{Updating criteria}\label{SECTION_UpdatingStrategies}
Defining a \mbox{SIMHO} strategy to find the optimal alternative within the available large budget is challenging. By using the information observed in the previous executions to update the distribution, the sampling can be efficiently guided towards it~\cite{mathesen_stochastic_2021,li_hyperband_2018}. This enables the efficient use of the multiple attempts available, allowing the \mbox{SIMHO} strategy to execute an alternative close to the highest quality. In practice, to update the distribution $p$ over $A$ using the observed information $\mathcal{I}$, it is sufficient that at least one component distribution $p_H,p_\Theta,p_\Xi$ or $p_C$ does so. The rest may not update, or if they do, may not require the use of the observed information for updating. In this sense, we classify the component distributions into three categories depending on the criteria used for their updating (summarized in Table~\ref{TABLE_FactorDistributionUpdate} and illustrated in {\color{mypurple}\textit{Rule~1}} of Figure~\ref{FIGURE_summary}):

\begin{table}[h]
{\footnotesize
\caption{Identification of the characteristics that distinguish the three updating criteria.}\label{TABLE_FactorDistributionUpdate}
\centering
\begin{tabular}{|c|cc|c|}
\hline 
\multirow{2}{*}{Component distribution}  &  \multicolumn{2}{c|}{Update type} & \multirow{2}{*}{Uses $\mathcal{I}$?} \\ 
  & Static & Dynamic &   \\ 
\hline 
\textit{Fixed}  & \checkmark &  &  \\ 
\hline 
\textit{Scheduled}  &  & \checkmark &  \\ 
\hline 
\textit{Adaptive}  &  & \checkmark & \checkmark\\ 
\hline 
\end{tabular} }
\end{table}

\textit{Fixed}. The component distribution is fixed when each component value is always assigned the same probability from iteration to iteration. Thus, the distribution remains static independently of how the observed information is extended (see the first column in {\color{mypurple}\textit{Rule~1}} of Figure~\ref{FIGURE_summary}). In the works where the set of algorithms $H$ or parameters $\Theta$ and initializations $\Xi$ define different alternatives~\cite{cicirello_max_2005,souravlias_parallel_2019,parsopoulos_parallel_2022,doerr_understanding_2020,auger_restart_2005,el-mihoub_multi-start_2022}, the distribution of initializations $p_\Xi$ is fixed. To extend the observed information of one of the available algorithms or parameters, it is executed again considering a randomly selected initialization, i.e., an initialization sampled from a static uniform distribution defined over $X$.

\textit{Scheduled}. Unlike a fixed distribution, a scheduled distribution is dynamic, although the observed information is still unused for its update. In this case, the different distributions to be considered in each iteration are predefined before executing any alternative (see the second column in {\color{mypurple}\textit{Rule~1}} of Figure~\ref{FIGURE_summary}). We observed this strategy for the stopping criterion distribution in some works where the set of parameters $\Theta$ or initializations $\Xi$ with stopping criteria $C$ define different alternatives. In each iteration, the same amount of resources are allocated for all non-discarded parameter configurations~\cite{li_hyperband_2018,das_dores_bandit-based_2018,falkner_bohb_2018,huang_asymptotically_2020} or initializations~\cite{jaderberg_population_2017}, which increases exponentially or linearly from iteration to iteration, respectively. Therefore, after defining the initial amount of resources, the criterion for all iterations is completely determined, i.e., it is the initial amount raised to the power of the number of iterations or with a proportional increase to iterations. Consequently, although $C$ is formed by all possible amounts of resources, without using $\mathcal{I}$ it is known that in each iteration $p_C$ assigns probability 1 to the amount inferred by the number of iterations and 0 to the rest. The same applies to the configuration distribution $p_\Theta$ in works focused on designing parameter-free evolutionary algorithms~\cite{auger_restart_2005,vrugt_self-adaptive_2009,doerr_understanding_2020}, where increasing population size strategies are proposed.

\textit{Adaptive}. As a scheduled distribution, an adaptive distribution is dynamic, but its updating is guided using the observed information. In the previous works where the set of parameters $\Theta$ and stopping criteria $C$ define different alternatives, in each iteration all the non-discarded parameter configurations are executed in parallel, i.e., it is known that all of them have been executed as many times as iterations elapsed, which allows predefining the stopping criterion distribution. However, there are also works that require the observed information to update this distribution. \citet{huang_efficient_2019} only execute one of the non-discarded configurations per iteration, so they consider individual geometric scheduling~\cite{provost_efficient_1999} for each of them. In this case, as the selected configuration for the current iteration has not been executed in all previous iterations, the number of times it has been executed can only be obtained from the history of configurations executed in each previous iteration, thus using $\mathcal{I}$. \citet{klein_fast_2017} instead, rather than conditioning $p_C$ by the set of discrete $\Theta$ parameter configurations, they treat this set continuously and define for all configurations the same discrete set of possible stopping criteria $C$ formed by different train set sizes. Finally, a Bayesian optimizer is applied that uses the observed information to determine at each iteration the configuration together with the most promising stopping criterion.

\subsection{Quality meaning of probability in adaptive criterion}\label{SECTION_ProbabilityMeaning}
The adaptive updating criterion is the only one that considers the observed information (see Table~\ref{TABLE_FactorDistributionUpdate}). Thus, unlike the fixed and scheduled criteria, the adaptive one allows to guide the sampling towards the best quality alternative efficiently. We have defined the probability of an unexecuted alternative as the preference with which the alternative will be sampled against the others. In practice, the adaptive criteria give to these preferences a meaning associated with the quality of the alternatives. Specifically, we distinguish two different meanings of the probabilities related to quality (see {\color{mypurple}\textit{Rule~2}} of Figure~\ref{FIGURE_summary}):

\begin{table}[h]
\caption{Classification of reviewed papers depending on the criterion used to update each component distribution and the quality meaning of probability derived from the adaptive ones. The classification of papers with at least one adaptive component is completed in the sections indicated in the last column and summarized in Tables~\ref{TABLE_UtilityPredictor} and~\ref{TABLE_UncertaintyControl}.}\label{TABLE_Overview}
\centering
{\footnotesize
\begin{tabular}{|c|cccc|cc|}
\hline
\multirow{2}{*}{Paper}  &  \multicolumn{4}{c|}{Updating criterion}& \multicolumn{2}{c|}{\makebox[2cm]{Quality meaning of $p$}} \\
  & $p_H$  & $p_\Theta$ & $p_\Xi$ & $p_C$ & {\footnotesize  \makebox[1.5cm]{\thead{EMQ\\(Section~\ref{SECTION_ProbabilityUpdateEMQ})}}} & {\footnotesize \makebox[1.5cm]{\thead{EQI\\(Section~\ref{SECTION_ProbabilityUpdateEQI})}}}\\
\hline 
\thead{\citep{addis_local_2005}\\\citep{siddiqui_investigating_2025}}   & - & - & \textit{Adaptive} & -&\checkmark &\\
\hline
\thead{\citep{luersen_globalized_2004}\\\citep{mathesen_stochastic_2021}\\\citep{meng_enhanced_2023} \\\citep{voglis_towards_2009}\\\citep{wang_history-guided_2023}  } & - & - & \textit{Adaptive} & - & &\checkmark\\
\hline 
\citep{thornton_auto-weka_2013}   & \textit{Adaptive}& \textit{Adaptive} & - & -  & & \checkmark\\
\hline 
\citep{hu_cascaded_2022} & \textit{Adaptive} & \thead{\textit{Fixed}\\\textit{Adaptive}} & - & -  & \checkmark & \checkmark\\
\hline
\thead{\citep{li_efficient_2020}\\\citep{schmidt_hamlet_2020}}  & \textit{Adaptive}& \textit{Adaptive} & - & -  &\checkmark & \checkmark\\
\hline 
\citep{cicirello_max_2005}    & \textit{Adaptive} &-& \textit{Fixed} & -&\checkmark &\\
\hline 
\thead{\citep{el-mihoub_multi-start_2022}} & \textit{Fixed} & - & \textit{Fixed} & -& &\\
\hline 
\thead{\citep{auger_restart_2005}\\\citep{doerr_understanding_2020}  } & - & \textit{Scheduled} & \textit{Fixed} & - & &\\
\hline 
\citep{vrugt_self-adaptive_2009} & - & \textit{Scheduled} & \textit{Adaptive} & - & \checkmark &\\
\hline 
\citep{huang_efficient_2019}  & -& \textit{Adaptive} & - & \textit{Adaptive} &\checkmark  &\\
\hline 
\citep{klein_fast_2017}  &  - & \textit{Adaptive} & - & \textit{Adaptive}& &\checkmark\\
\hline
 \citep{li_hyperband_2018}   & - & \textit{Adaptive} & - & \textit{Scheduled} & \checkmark &\\
\hline 
 \thead{\citep{falkner_bohb_2018}\\\citep{huang_asymptotically_2020}}    & - & \textit{Adaptive} & - & \textit{Scheduled} & \checkmark & \checkmark\\
\hline 
\citep{gyorgy_efficient_2011}  &- & - & \textit{Adaptive} & \textit{Adaptive}& \checkmark &\\
\hline
\citep{souravlias_parallel_2019}    & \textit{Adaptive} &  \textit{Adaptive}&\textit{Fixed} & -&\checkmark &\\
\hline 
\citep{das_dores_bandit-based_2018}   &\textit{Adaptive} & \textit{Adaptive} & - & \textit{Scheduled}&\checkmark &\\
\hline 
\citep{parsopoulos_parallel_2022}    & \textit{Fixed} & - &\textit{Fixed} & \textit{Adaptive}& \checkmark &\\
\hline 
\citep{jaderberg_population_2017}   & - & \textit{Fixed} &\textit{Adaptive} & \textit{Scheduled}&\checkmark &\\
\hline
\end{tabular}}
\end{table}

\textit{Expected maximum quality} (EMQ). In some papers, higher probabilities are associated with the alternatives expected to have the highest quality considering the observed information. Therefore, in each iteration, the alternative with the highest predicted quality is preferred. For example, \citet{souravlias_parallel_2019} assign more CPUs to the pairs of algorithms and parameter configurations with higher average qualities observed in their previous executions with different initializations. Thus, the probability of the alternatives with the pair of algorithm and configuration of higher estimated quality is greater. 

\textit{Expected quality improvement} (EQI). In other papers instead, higher probabilities are assigned to alternatives that are expected to improve on the highest quality observed so far among the executed alternatives. Consequently, the improvement in quality is followed to reach the maximum.  A clear example is given by \citet{luersen_globalized_2004}. They assign higher probabilities to initializations that are believed to yield the exploration of new local optima not visited by the Nelder-Mead algorithm in its previous executions. In this context, exploring a region of the solution space associated with a different local optimum is desirable, as it is more likely to improve the best-observed solution.

We have only referenced some papers throughout the previous  Section~\ref{SECTION_UpdatingStrategies} and the current section to illustrate the proposed categories. However, in Table~\ref{TABLE_Overview}, we classify all the papers considered in this review according to these categories. 

\subsection{Two-stage process for adaptive criterion}
The dependence of the adaptive criterion on the observed information requires knowing how to use it to define the update of the distribution iteratively. In the literature, we identify a two-stage process for the updating of an adaptive component distribution based on the observed information (see {\color{mypurple}\textit{Rule~3}} of Figure~\ref{FIGURE_summary}):

\textit{Stage 1: Prediction of utility.} Firstly, to assign a probability to each value of the component, its utility is computed, i.e., the observed information is used to predict how useful the knowledge about the information of a new alternative with that component value would be in the search for the highest quality alternative. Depending on which of the three variables forming the observed information $\mathcal{I}$ are used, i.e., the best-visited solutions $x_a$, the execution times $t_a$ or the qualities $f_a$ of executed alternatives $a\in A$, the utility prediction takes different meanings that are described in the following two sections (summarized in the Table~\ref{TABLE_UtilityPredictor}).

\textit{Stage 2: Control of uncertainty}. Secondly, the uncertainty of the predicted utilities must be controlled to determine the final component distribution. When the number of executed alternatives is low, the representation of the total information is poor, leading to uncertain utility predictions. In this case, the probabilities should be less biased by the utilities to avoid ignoring relatively different alternatives with promising information. Consequently, as the amount of the observed information increases, the representation of the total information improves, reducing the uncertainty of the utilities (see the third column in {\color{mypurple}\textit{Rule~1}} together with {\color{mypurple}\textit{Rule~3}} of Figure~\ref{FIGURE_summary}). Therefore, the uncertainty should be controlled considering the resources consumed so far by the \mbox{SIMHO} strategy, i.e., the execution time elapsed or the number of iterations spent in obtaining the observed information. This can be carried out in an \textit{explicit} or \textit{implicit} manner, depending on whether the resources consumed interact analytically with the utility in defining the probability or not, respectively. In the remainder of this section, we describe the different classifications that are given in practice (summarized in Table~\ref{TABLE_UncertaintyControl}). 

The following sections exemplify the two-stage process depending on the meaning to be considered for the probability. In each section, we describe how the utility is predicted and how its uncertainty is controlled in practice (possible settings of Table~\ref{TABLE_SummaryAdaptiveFactorDistributionUpdate}).
\begin{table}[h]
{\footnotesize
\caption{Characteristics to be considered for the update of an adaptive component distribution.}\label{TABLE_SummaryAdaptiveFactorDistributionUpdate}
\centering
\begin{tabular}{|c|c|}
\hline 
\multicolumn{2}{|c|}{\textit{Adaptive} component distribution} \\ 
\hline 
How to predict utility? & How to control uncertainty? \\ 
\hline 
\thead{Determining the use of\\the observed information\\[2mm] $\mathcal{I}=\big\lbrace \hspace{-2mm}\underbrace{(x_a,t_a,f_a)}_{\textstyle\textit{Which to use?}}\hspace{-2mm}~|~ a \text{ is executed}\big\rbrace$} 
&
\thead{Considering \\[1mm] \textit{Explicit}/ \textit{Implicit} \\[1mm]
interaction of consumed resources \\ with utility to define probability}
 \\ 
\hline
\end{tabular} }
\end{table}

\subsubsection{Probability related to expected maximum quality}\label{SECTION_ProbabilityUpdateEMQ}
One of the options for guiding sampling toward the highest-quality alternative is to assign higher probabilities to the component values expected to define the highest-quality alternatives. In the papers considering this association of quality with a component distribution, we observed that the utility predictors are based only on the qualities of the observed information (the $f_a\in\mathcal{I}$) or these are combined with their execution times (the $t_a\in\mathcal{I}$). Moreover, the uncertainty of the predicted utilities is measured either explicitly or implicitly.

\begin{table}[h]
\caption{Classification of reviewed papers that address the selection of at least one adaptive component depending on the meaning of the utility predictor and the observed information used for it. Papers are grouped by common utility predictor and ordered following their description in Sections~\ref{SECTION_ProbabilityUpdateEMQ} and~\ref{SECTION_ProbabilityUpdateEQI}. A paper may appear multiple times when different utility predictors are considered for updating the adaptive components shown in Table~\ref{TABLE_Overview}.}\label{TABLE_UtilityPredictor}
\vspace{-2mm}
\setlength{\tabcolsep}{13pt}
\centering
{\footnotesize
\begin{tabular}{|c|c|ccc|}
\hline 
\multirow{2}{*}{Paper} & \multirow{2}{*}{Utility predictor} & \multicolumn{3}{c|}{\makebox[0pt]{Which to use from $\mathcal{I}$?}} \\ 
 &  & $x_a$ & $t_a$ & $f_a$ \\ 
\hline 
\thead{\citep{das_dores_bandit-based_2018}\\\citep{falkner_bohb_2018}\\\citep{gyorgy_efficient_2011}\\\citep{huang_asymptotically_2020}\\\citep{jaderberg_population_2017}\\\citep{li_hyperband_2018}\\\citep{parsopoulos_parallel_2022}\\\citep{siddiqui_investigating_2025}\\\citep{souravlias_parallel_2019}\\\citep{vrugt_self-adaptive_2009}} & \multirow{2}{*}{Point estimation} &  &  & \checkmark \\
\cline{1-1}\cline{3-5} 
\citep{hu_cascaded_2022} &  &  & \checkmark & \checkmark \\ 
\hline 
\thead{\citep{addis_local_2005}\\\citep{cicirello_max_2005}\\\citep{falkner_bohb_2018}\\\citep{huang_asymptotically_2020}\\\citep{klein_fast_2017}\\\citep{mathesen_stochastic_2021}\\\citep{meng_enhanced_2023}} & Density estimation &  &  & \checkmark \\ 
\hline 
\thead{\citep{huang_efficient_2019}\\\citep{li_efficient_2020}} & \multirow{2}{*}{Interval estimation} &  & \checkmark & \checkmark \\ 
\cline{1-1}\cline{3-5} 
\citep{thornton_auto-weka_2013} &  &  &  & \checkmark \\ 
\hline 
\thead{\citep{hu_cascaded_2022}\\\citep{schmidt_hamlet_2020}} & Extrapolation &  & \checkmark & \checkmark \\ 
\hline 
\thead{\citep{luersen_globalized_2004}\\\citep{voglis_towards_2009}\\\citep{wang_history-guided_2023}} & Exploratory probability & \checkmark &  &  \\ 
\hline 
\end{tabular} }
\end{table}

\begin{itemize}
\item \textit{Utility prediction with $f_a$}. Among the works that only use the qualities, we find those that estimate the qualities in a pointwise manner, calculating weighted averages of the $f_a$ observed in the previous executions of each algorithm~\cite{souravlias_parallel_2019,parsopoulos_parallel_2022}, or using their maxima as an estimate of the quality of a parameter configuration~\cite{li_hyperband_2018,falkner_bohb_2018,huang_asymptotically_2020} or an initialization~\cite{gyorgy_efficient_2011,vrugt_self-adaptive_2009,siddiqui_investigating_2025}. Other works instead, carry out density estimations. \citet{cicirello_max_2005} first estimate the Gumbel distribution of the maximum quality of each algorithm, and then calculate the expected value using that distribution. \citet{addis_local_2005} determine the restart position for the local search using a smooth approximation of the objective function. They define a local Gaussian kernel estimate of the piecewise constant function represented by the local optima and its objective values. Since only the observed local optima from the executed initializations are known, the utility of a new initialization is approximated using the kernel estimate restricted to the qualities of the nearby explored starting points.

\item \textit{Utility prediction with $f_a$ and $t_a$}. Regarding the works that also use the execution times, we still find point estimators that use the times in addition to the qualities to calculate averages~\cite{hu_cascaded_2022}. However, we also find interval estimates. They consist of defining an upper and lower bound of the quality of algorithms~\cite{li_efficient_2020} or parameter configurations~\cite{huang_efficient_2019} using the $f_a$ and $t_a$ observed for each of them. Moreover, we found works with more elaborate proposals, which try to model the learning curve~\cite{schmidt_hamlet_2020} or convergence curve~\cite{hu_cascaded_2022} of the hyperparameter optimizer of an algorithm, to consider its extrapolation in the remaining execution time as the utility prediction of each algorithm.

\item \textit{Explicit uncertainty control}. Among the techniques with explicit control of uncertainty, we find Boltzmann exploration~\cite{cicirello_max_2005} and decay $\varepsilon$-greedy~\cite{schmidt_hamlet_2020}. Their respective temperature parameter and $\varepsilon$ decrease with the time or iterations elapsed in all previous executions of the algorithms, allowing the probabilities assigned to each algorithm to get closer to their predicted utilities over iterations. Moreover, methods based on upper confidence bounds (UCB) explicitly add to each utility an amount inversely proportional to the time or iterations spent in each algorithm~\cite{schmidt_hamlet_2020,hu_cascaded_2022} or initialization~\cite{gyorgy_efficient_2011}. Therefore, higher utilities of algorithms or initializations in which more resources are invested are equalled to smaller predicted utilities for those with less investment.

\item \textit{Implicit uncertainty control}. For implicit control of uncertainty instead, we find techniques that are based on discarding. At each iteration the algorithms, parameter configurations or initializations that have a significantly worse utility than the rest are discarded, e.g., an estimated quality interval with an upper bound below the lower bound of the rest~\cite{li_efficient_2020} or a maximum quality value below the top percentage in the ranking~\cite{li_hyperband_2018,das_dores_bandit-based_2018,jaderberg_population_2017,falkner_bohb_2018,huang_asymptotically_2020}. Notice that in these cases, the number of algorithms, configurations or initializations not discarded that will be executed again in the next iteration decreases with the growth of iterations, increasingly allocating more resources of an iteration to those with higher predicted utilities. In addition, \citet{huang_efficient_2019}, after discarding configurations by comparing their estimated confidence intervals of quality, compare the interval shortening rates of the rest to select a unique configuration to be executed in the next iteration. 

\item[]In contrast, \citet{souravlias_parallel_2019} allocate the number of available CPUs among the different algorithms proportionally to their pointwise estimates of quality. \citet{parsopoulos_parallel_2022} instead, dedicate each worker node to a portfolio algorithm and use an adaptive pursuit strategy to allocate probabilistically the resources invested in each parallel batch based on the point estimates of quality of each algorithm in previous batches. Therefore, without discarding any algorithm, the number of iterations spent in all of them regulates the uncertainty of the utilities, i.e., the estimates become more certain with iterations, allocating more resources to the most promising algorithms. Moreover, \citet{schmidt_hamlet_2020} consider the $\varepsilon$-greedy technique in which $\varepsilon$ is the invariant preference defined at all iterations for all algorithms that are not the most useful. Consequently, the uncertainty of their utilities is measured in a constant manner, without discarding any algorithm for having assigned as minimum preference $\varepsilon$.

Other works related to evolutionary computation instead, consider random sampling techniques to control the uncertainty associated with the initialization of the population. They randomly initialize population-based algorithms resulting from hyper-heuristics~\cite{vrugt_self-adaptive_2009} or quantum-inspired evolutionary algorithms~\cite{siddiqui_investigating_2025}, while incorporating a subset of the previously visited best solutions. These strategies aim to strike a balance between retaining promising solutions and exploring new regions of the search space. Thus, uncertainty is controlled by that portion of the randomly initiated population.

\end{itemize}

\begin{table}[h]
\caption{Classification of reviewed papers that address the selection of at least one adaptive component depending on how they control the uncertainty of utilities. Papers are grouped by common uncertainty controller and ordered following their description in Sections~\ref{SECTION_ProbabilityUpdateEMQ} and~\ref{SECTION_ProbabilityUpdateEQI}. A paper may appear multiple
times when different uncertainty controllers are considered for updating the adaptive components shown in Table~\ref{TABLE_Overview}.}\label{TABLE_UncertaintyControl}

\setlength{\tabcolsep}{10pt}
\centering
{\footnotesize
\begin{tabular}{|c|c|cc|}
\hline 
\multirow{2}{*}{Paper} & \multirow{2}{*}{Uncertainty controller} & \multicolumn{2}{c|}{How is the control?}    \\ 
 &  & \textit{Explicit} & \textit{Implicit}\\ 
\hline 
\citep{schmidt_hamlet_2020} & Decay $\varepsilon$-greedy  & \checkmark &  \\  
\hline 
\thead{\citep{gyorgy_efficient_2011}\\\citep{hu_cascaded_2022}\\\citep{schmidt_hamlet_2020}} & UCB + argmax  & \checkmark &      \\
\hline 
\thead{\citep{das_dores_bandit-based_2018}\\\citep{falkner_bohb_2018}\\\citep{huang_asymptotically_2020}\\\citep{li_hyperband_2018}} & Successive discarding &  &\checkmark \\  
\hline 
\thead{\citep{huang_efficient_2019}\\\citep{li_efficient_2020}} & Lower-upper bound-based &  & \checkmark \\
\hline 
\citep{souravlias_parallel_2019} & CPU allocation &  & \checkmark  \\  
\hline 
\citep{parsopoulos_parallel_2022} & Adaptive pursuit resource allocation &  & \checkmark  \\  
\hline 
\citep{schmidt_hamlet_2020} & $\varepsilon$-greedy  &  & \checkmark  \\ 
\hline 
\thead{\citep{falkner_bohb_2018}\\\citep{huang_asymptotically_2020}\\\citep{klein_fast_2017}\\\citep{mathesen_stochastic_2021}\\\citep{meng_enhanced_2023}\\\citep{thornton_auto-weka_2013}} & Acquisition function + argmax  &  & \checkmark   \\
\hline 
\thead{\citep{addis_local_2005}\\\citep{jaderberg_population_2017}\\\citep{luersen_globalized_2004}\\\citep{siddiqui_investigating_2025}\\\citep{voglis_towards_2009}\\\citep{vrugt_self-adaptive_2009}\\\citep{wang_history-guided_2023}} & Random sampling  &  & \checkmark  \\ 
\hline 
\citep{cicirello_max_2005} & Boltzmann exploration & \checkmark & \\ 
\hline 

\end{tabular} }
\end{table}

\subsubsection{Probability related to expected quality improvement}\label{SECTION_ProbabilityUpdateEQI}

The second option for guiding sampling toward the optimal alternative consists of associating higher probabilities to the component values expected to define alternatives with more chances for improving the best observed quality. For a component distribution that defines probabilities with this meaning, we found works that continue defining the utility predictor considering the qualities of the observed information (the $f_a\in\mathcal{I}$), although we also observed the use of only the best-visited solutions to predict utilities (the $x_a\in\mathcal{I}$). However, in all of them, an implicit control of uncertainties is always considered. 

\begin{itemize}
\item \textit{Utility prediction with $f_a$}. The literature techniques that predict the utilities using the qualities correspond to the first stage of Bayesian optimization, i.e., the prediction by a model. Utility predictors are models that capture the dependence between the component value and the $f_a$ of the alternatives they define. The Gaussian process is the most commonly used surrogate model, which estimates a distribution for the quality of each unexecuted parameter configuration~\cite{klein_fast_2017,hu_cascaded_2022} or initialization~\cite{mathesen_stochastic_2021,meng_enhanced_2023} or stopping criterion~\cite{klein_fast_2017}, after training the model with the pairs formed by the executed component values and their observed qualities. \citet{thornton_auto-weka_2013} instead, use random forest models to predict simultaneously the quality mean and variance of algorithm and parameter configuration combinations, arguing that they are good predictors with discrete and high-dimensional input data. In other works~\cite{falkner_bohb_2018,huang_asymptotically_2020}, an adaptation of the Tree Parzen Estimator~\cite{bergstra_algorithms_2011} is used to model densities over the executed parameter configuration space.

\item \textit{Utility prediction with $x_a$}. As an example of utility prediction that uses the best-visited solutions, \citet{luersen_globalized_2004} estimate the density of the set formed by the pairs of starting points $\xi$ and local convergence points $x_a$ visited in each previous execution of the Nelder-Mead algorithm. They argue that the topology of the basins of attraction can be efficiently summarized by using only these two points among all the points visited during the local searches. The utility predictor is the density estimator obtained by applying a Gaussian Parzen-Window. Thus, the predicted utility for each new initialization is the probability that the Nelder-Mead departed from that point can explore a different region of the solution space and converge to a new local optimum. Similarly, \citet{voglis_towards_2009} estimate the probability that an initial point does not belong to the region of attraction of any of the previously identified minima. These probabilities depend on the distances between the initial points and the local minima found by the Broyden–Fletcher–Goldfarb–Shanno (BFGS) algorithm~\cite{rinnooy_kan_stochastic_1987}, the direction of the gradient, and the frequency with which each minimum has been discovered. \citet{wang_history-guided_2023} consider a similar potential metric for evolutionary computation methods. They show how the populations of evolutionary methods with simple restarts tend to repeat previous evolution processes ending in the basins of attraction already explored. To address this issue, they use the evaluated solutions to divide the search space into hill regions for estimating the basins of attractions, and encourage further exploration in those of higher potential for finding unidentified ones.

\item \textit{Implicit uncertainty control}. Although we only found techniques that control the uncertainty of the utilities implicitly, we distinguish some that consider acquisition functions and others that apply random samplings. 

When using a Bayesian optimization, the uncertainty of the quality distributions predicted by the model for each component value is controlled by an acquisition function. Among the most popular options is the expected improvement. It maps to each initialization~\citep{mathesen_stochastic_2021,meng_enhanced_2023} or parameter configuration~\citep{falkner_bohb_2018,huang_asymptotically_2020} the expected quality increment over the best observed, which is larger for values predicted with higher mean or variance of quality. \citet{klein_fast_2017} also use the more recent entropy search acquisition function. It favours component values with larger expected information gains about the optimum, rather than trying to evaluate close to the optimum. Notice that the larger the observed information, the predicted distributions have a lower variance, leading to acquisition functions closer to the predicted model. Thus, the number of iterations implicitly measures the uncertainty.

Other works instead, use a random sampling technique to control the uncertainty of the predicted exploratory probabilities. \citet{luersen_globalized_2004} randomly select a subset of unexecuted initializations, and then choose the one with the highest predicted probability among them to reinitialize the algorithm from that location. \citet{voglis_towards_2009} uniformly sample starting points until they find one with a probability of not belonging to a region of attraction that is greater than a randomly selected number from a uniform distribution between 0 and 1. The first starting point that meets this condition is used for the next execution of the BFGS algorithm. Therefore, these techniques allow selecting initializations that are not the best according to the predicted utilities, although it is still guided by them. Moreover, as the considered samplings are always random, the uncertainty of the predicted probabilities is measured constantly.
\end{itemize}

\section{Discussion}\label{SECTION_Discussion}

This section discusses the implications of the proposed framework and the scope of the review. First, we reflect on the utility of the framework as a conceptual tool to organize existing single-problem selection strategies and to support the design of new ones. Second, we discuss the relationship between single- and multi-problem settings, with a particular focus on the transferability of strategies across these settings and on the intended scope and limitations of this review.

\subsection{Framework utility and design of new strategies}
In the previous section, we characterized the valid strategies for SIMHO within the proposed framework. This review is summarized in Table~\ref{TABLE_Overview}, Table~\ref{TABLE_UtilityPredictor} and Table~\ref{TABLE_UncertaintyControl}. The combined use of these three tables enables practitioners to identify, choose and design appropriate SIMHO strategies by navigating through the components of the framework. Given the flexibility of the proposed framework, this review can be naturally extended by incorporating in the summary tables other selection strategies related to the literature topics of algorithm selection, parameter tuning, multi-start, and resource allocation.

The tables provide easy access to selection strategies associated with different components. For example, a reader interested in multi-start (MS) strategies can identify relevant works in Table~\ref{TABLE_Overview} by checking which rows are filled in the $p_\Xi$ column. The corresponding section for each adaptive strategy can then be located using the last columns of Table~\ref{TABLE_Overview}, Table~\ref{TABLE_UtilityPredictor}, and Table~\ref{TABLE_UncertaintyControl}, which indicate where each strategy is discussed. Specifically, papers that have a checkmark in EMQ or EQI of the last column of Table~\ref{TABLE_Overview} are described in one of the items in Sections~\ref{SECTION_ProbabilityUpdateEMQ} or~\ref{SECTION_ProbabilityUpdateEQI}, respectively, and this item is completely determined by the checkmarks in the last columns of Table~\ref{TABLE_UtilityPredictor} and Table~\ref{TABLE_UncertaintyControl} associated with that paper. Based on this information, the reader can gain a general understanding of the main characteristics of each strategy and choose the one that best aligns with their specific requirements and preferences. Then, the reader may identify several MS strategies suitable for their problem, such as randomized restarts~\cite{cicirello_max_2005}, Bayesian optimization-based restarts~\cite{mathesen_stochastic_2021}, or kernel density estimation-based strategies~\cite{luersen_globalized_2004}, and gain a detailed understanding of each approach by consulting the corresponding references in the literature. Moreover, since an alternative is defined as any combination of the four components, some readers may also be interested in strategies that address the selection of multiple components simultaneously. For instance, a reader seeking both parameter tuning (PT) and multi-start (MS) may combine a configuration selection strategy from one work with an initialization selection strategy from another. Alternatively, the reader may adapt strategies originally developed for one component to reuse them in the selection of a different component of interest. Indeed, we observe that the same strategies are often applied to different components across various works. For example, Bayesian optimization is employed for both configuration~\cite{klein_fast_2017,hu_cascaded_2022} and initialization~\cite{mathesen_stochastic_2021,meng_enhanced_2023}, while exponential-increasing strategies are associated with both configuration~\cite{auger_restart_2005,doerr_understanding_2020} and stopping criteria~\cite{huang_asymptotically_2020,li_hyperband_2018}.

The last two utilities described for the review, namely combination and reuse, can lead to the design of new strategies, which would automatically be classified within our framework. Therefore, any newly designed strategy that can be characterized by our taxonomy is considered a valid strategy for addressing SIMHO. As a result, the proposed taxonomy facilitates both the development and identification of suitable strategies, including all new or existing approaches that fall within its structure. Furthermore, the \textit{Updating criterion} column in Table~\ref{TABLE_Overview} allows us to identify underexplored types of strategies and guide future research to address these gaps. Our review shows that 84\% of the included papers propose strategies for selecting one or two components, whereas only 16\% address the selection of three components. This indicates that new and improved strategies could be those that consider the selection of more components, or even all four, simultaneously, as shown in recent experimental analyses~\cite{el-mihoub_multi-start_2022}.

\subsection{Transferability of strategies across settings and scope of the review}

From a purely formal perspective, one could argue that the single-problem setting may be seen as a particular variant of the multi-problem setting in which only one instance is available (see Appendix~\ref{APPENDIX_MultiproblemSetting} for a more detailed discussion). However, this formal inclusion does not imply methodological equivalence or transferability.

Strategies designed for multi-problem settings are inherently built around the availability of multiple instances and the possibility of transfer or generalization across them. When these assumptions do not hold, as in the single-problem setting, such strategies either cannot be applied effectively or change their objective and interpretation. For example, per-instance algorithm selection~\citep{rice_algorithm_1976} relies on supervised models trained on features extracted from a set of problem instances. When only a single problem is available, such models cannot be constructed efficiently due to the lack of training instances or sufficiently related problems. In contrast, methods such as irace~\citep{lopez-ibanez_irace_2016}, which are designed to identify configurations that perform robustly across multiple instances, can be adapted to the single-problem setting by repeatedly racing configurations on the same problem~\citep{bischl_hyperparameter_2023}. In this case, the strategy effectively shifts into a resource allocation or successive halving scheme, rather than a method for identifying configurations that are robust across instances. This is conceptually similar to approaches such as Hyperband~\citep{li_hyperband_2018}, which are included in our review.

Whether the single-problem setting should be considered a niche therefore strongly depends on the target audience. For practitioners and researchers interested in single-problem optimization, existing reviews leave a clear gap: they either study strategies originally designed for this setting separately or include multi-problem strategies that are not efficiently transferable to the single-problem case. By unifying and systematizing the strategies that were originally designed for the single-problem setting, our review directly addresses this gap and provides a focused and coherent reference for this audience.

Conversely, our review is not intended for users interested in multi-problem optimization, as it is intentionally and exclusively focused on the single-problem setting. Moreover, single-problem strategies are generally not efficiently transferable to the multi-problem case either. For instance, such strategies could be applied in a plug-and-play manner to each instance independently, but this would ignore cross-instance information and therefore fail to exploit robustness or transfer. Similarly, in multi-start strategies~\citep{luersen_globalized_2004,mathesen_stochastic_2021} that progressively learn the most interesting regions of the search space to start a local search, changing the problem would not be meaningful, because the learned information that is used to guide new local search runs is specific to the problem instance and its search space.

Overall, this review is intentionally and exclusively focused on the single-problem setting, motivated by its practical relevance and by the lack of a unified and systematic review dedicated to it. As a consequence of this deliberate focus, the review does not aim to address the needs of practitioners interested in multi-problem optimization. That setting is already covered by existing surveys, some of which are listed in Table~\ref{TABLE_contribution}, although evaluating or recommending them falls outside the scope of this work.

\section{Conclusion}\label{SECTION_Conclusion}
This paper presents a focused review of SIMHO, which refers to the scenario of single-problem multi-attempt heuristic optimization. In this scenario, a practitioner seeks to find the best possible solution to a specific optimization problem, relying on heuristic methods and using only the information gathered throughout its optimization processes. Heuristic methods are often preferred in practice due to the challenging nature of certain optimization problems, which frequently involve objective functions that are large in scale, non-convex, or affected by noise. The performance of heuristic methods depends on four key components: the algorithm, its parameter configuration, the initialization, and the stopping criterion. The execution of a different combination of these components, known as an alternative, leads to a different optimization process of the same given optimization problem. When the available computational budget is sufficiently large, multiple attempts are allowed, making it possible to execute a different alternative in each one. In such cases, the sequential selection of alternatives plays a critical role in efficiently identifying the best combination of the four components and, ultimately, in finding a high-quality solution to the optimization problem.

Although SIMHO is commonly encountered in practice, to the best of our knowledge, this is the first review that focuses exclusively on this problem and unifies alternative selection strategies that address the selection across all four components. Selection strategies suitable for SIMHO have been studied separately within the topics of algorithm selection, parameter tuning, multi-start, and resource allocation. Existing reviews fail to provide a comprehensive view, either because they do not consider strategies from all four topics or because they include others that are not suitable for SIMHO scenario. As a result, practitioners aiming to address SIMHO face limitations in understanding and applying the broad range of existing strategies. This review addresses that gap by bringing together strategies that, although developed independently, are all suitable for tackling SIMHO. We formalize these strategies within a unified framework using consistent terminology, and introduce a taxonomy that organizes and characterizes them accordingly. This framework-based review thus facilitates both the identification and the development of selection strategies for SIMHO.

The framework proposed in this review unifies and conceptually describes the existing alternative selection strategies as equally valid for SIMHO. This conceptual framework could serve as the foundation for a dedicated software library designed to address SIMHO. Such a library would enable the integration and combination of diverse selection strategies from the literature for targeting different components, and maybe would include options for comparative performance analysis on different problems. This tool could fulfill a role similar to that of HyFlex~\cite{hutchison_hyflex_2012}, which has supported the development of improved hyper-heuristics~\cite{alanazi_adaptive_2016,drake_recent_2020,ferreira_multi-armed_2017,ferreira_multi-armed_2015,sabar_dynamic_2015}.

\bibliographystyle{ACM-Reference-Format}
\bibliography{main.bib}

\appendix

\section{Hierarchical dependence between components}\label{APPENDIX_HierarchicalDep}

In practice, there is a hierarchical dependency between heuristic algorithms and their configurations, initializations and stopping criteria. For example, two different algorithms, such as the Nelder-Mead~\cite{nelder_simplex_1965} direct search algorithm or the evolutionary Genetic Algorithms~\cite{glover_handbook_2003}, do not share the same spaces of the last three components. In the first one, the reflection, expansion, contraction and reduction coefficients of the simplex are part of the configuration, the initialization is a single solution of the solution space and the stopping criteria are related to the shape and size of a simplex. For the second one, these components do not apply. Among the configuration are the selection, crossover and mutation operators of a population, the initialization is a set of solutions that forms the initial population, and the stopping criteria can be associated with the statistics in the population. Therefore, to be exact, we should denote $\Theta_h$, $\Xi_h$ and $C_h$, but for simplicity, we omit the subindex by implicitly assuming that $\Theta=\cup_{h\in H} \Theta_h$ and so on.

This dependence is also implicit in Definition~\eqref{DEFINITION_Probability} of the probability distribution of alternatives. We express it as a chain of conditional distributions, where the choice of each component is conditioned on those selected before it. For each $h\in H$ the support of the component distributions $p_\Theta$, $p_\Xi$ and $p_C$ is the subset $\Theta_h$, $\Xi_h$ and $C_h$, respectively. For example, it is always satisfied that $p_\Theta(\theta \mid h) = 0$ for all $\theta\notin \Theta_h$, while $\sum_{\theta\in\Theta_h}p_\Theta(\theta \mid h) = 1$.

\section{Formal similarity between single- and multi-problem settings}\label{APPENDIX_MultiproblemSetting}
Problem~\eqref{EQUATION_AOproblem} provides a unified formalization of the optimization problem underlying the single-problem multi-attempt setting considered in this review. Since this paper is a review focused entirely on the single-problem setting, the main role of the formalization, in Section~\ref{SECTION_Framework}, is to establish a shared terminology that allows us to systematically describe, in Section~\ref{SECTION_Taxonomy_UpdateDistribution}, selection strategies that were originally studied separately but are all suitable for this setting. Therefore, this formalization, which serves as a supporting tool to make this review more structured and coherent, is defined specifically for the single-problem setting and does not directly extend to the multi-problem setting, despite possible formal similarities that may be observed. It is important to emphasize that the framework is also not intended to model the multi-problem setting, since from the outset the main motivation of this review is the exclusive focus on SIMHO due to its practical relevance, as described in Section~\ref{SECTION_Introduction}.

Specifically, in our framework, Problem~\eqref{EQUATION_AOproblem} formalizes the optimization objective of identifying the alternative $a\in A$ that yields the best performance for a single optimization problem defined by a given objective function $f$ and search space $X$, i.e., an objective of the form 
$\argmax_{a\in A}~f_a$. In contrast, optimization problems in the multi-problem setting involve fundamentally different objectives that are defined over collections of problem instances rather than a single one, as discussed in Section~\ref{SECTION_RelatedWork}. For example, in per-instance algorithm selection~\citep{rice_algorithm_1976}, the objective is to identify, for each problem instance, the algorithm that performs best. The learned mapping from instances to algorithms is the strategy used to achieve this goal, but the underlying objective would correspond to an extension of Problem~\eqref{EQUATION_AOproblem} over a set of objective functions (i.e., a “for all” over multiple $f$’s), which is not captured by our formulation. Similarly, methods such as irace~\citep{lopez-ibanez_irace_2016} aim to identify configurations that perform robustly across a collection of instances, which would require a multi-objective or aggregated optimization objective incorporating performance over all instances of interest. In both cases, the availability of multiple problem instances is an explicit assumption, which does not align with the Problem~\eqref{EQUATION_OptimizaionProblem} underlying Problem~\eqref{EQUATION_AOproblem}.

\end{document}